\documentclass[sigconf]{acmart}
\usepackage{xspace}
\usepackage{booktabs} 
\usepackage{hyperref} 
\usepackage{graphicx}

\newcommand{\authorwithoutsuperscript}{Chao-Chun Hsu, Yu-Hua Chen, Zi-Yuan Chen, Hsin-Yu Lin, Ting-Hao (Kenneth) Huang, Lun-Wei Ku\xspace}

\newcommand{\system}{Dixit\xspace}

\newcommand\blfootnote[1]{%
  \begingroup
  \renewcommand\thefootnote{}\footnote{#1}%
  \addtocounter{footnote}{-1}%
  \endgroup
}

\copyrightyear{2019}
\acmYear{2019}
\setcopyright{iw3c2w3}
\acmConference[WWW '19]{Proceedings of the 2019 World Wide Web Conference}{May 13--17, 2019}{San Francisco, CA, USA}
\acmBooktitle{Proceedings of the 2019 World Wide Web Conference (WWW '19), May 13--17, 2019, San Francisco, CA, USA}
\acmPrice{}
\acmDOI{10.1145/3308558.3314131}
\acmISBN{978-1-4503-6674-8/19/05}

\usepackage{balance}

\begin{document}
\title{\system: Interactive Visual Storytelling via Term Manipulation}

\author{Chao-Chun Hsu$^{*}$}
\affiliation{%
   \institution{Academia Sinica}
   \city{Taipei}
   \country{Taiwan}}
\email{joe32140@iis.sinica.edu.tw}

\author{Yu-Hua Chen$^{*}$}
\affiliation{%
   \institution{Academia Sinica}
   \city{Taipei}
   \country{Taiwan}}
\email{cloud60138@iis.sinica.edu.tw}

\author{Zi-Yuan Chen$^{*}$}
\affiliation{%
   \institution{Academia Sinica}
   \city{Taipei}
   \country{Taiwan}}
\email{zychen@iis.sinica.edu.tw}

\author{Hsin-Yu Lin}
\affiliation{%
   \institution{Academia Sinica}
   \city{Taipei}
   \country{Taiwan}}
\email{swat230@iis.sinica.edu.tw}

\author{Ting-Hao (Kenneth) Huang}
\affiliation{%
   \institution{Pennsylvania State University}
   \city{State College}
   \state{PA}
   \country{USA}}
\email{txh710@psu.edu}

\author{Lun-Wei Ku}
\affiliation{%
   \institution{Academia Sinica}
   \institution{Most Joint Research Center for AI Technology and All Vista Healthcare}
   \city{Taipei}
   \country{Taiwan}}
\email{lwku@iis.sinica.edu.tw}


\begin{abstract}
In this paper, we introduce \textbf{\system}, an interactive visual
storytelling system that the user interacts with iteratively to compose a short
story for a photo sequence.
The user initiates the process by uploading a sequence of photos.
\system first extracts text \textit{terms} from each photo which describe the
objects (e.g., boy, bike) or actions (e.g., sleep) in the photo,
and then allows the user to add new terms or remove existing terms.
\system then generates a short story based on these terms.
Behind the scenes, \system uses an LSTM-based model trained on image caption
data and FrameNet to distill terms from each image, and utilizes a transformer
decoder to compose a context-coherent story.
Users change images or terms iteratively with \system to create
the most ideal story. \system also allows users to manually edit and rate
stories.
The proposed procedure opens up possibilities for \textit{interpretable} and
\textit{controllable} visual storytelling, allowing users to understand the
story formation rationale and to intervene in the generation process.

\end{abstract}

%



\maketitle
\blfootnote{$*$ denotes equal contribution}
\section{Introduction}
In visual storytelling, as introduced by Huang et
al.~\cite{huang2016visual}, we take a sequence of photos as input and attempt to
generate a short story narrating the sequence. 
This is an interdisciplinary task comprising image semantic reasoning and
structural text generation. 
In contrast to image captioning or video captioning, which have less
annotation variation across annotators, authors of visual stories can set
down totally different stories for a given image sequence based on their
preference and imagination.
Instead of generating one story and leaving it \textit{to} the user, we believe
that working on a story draft together and iterating \textit{with} the user
results in much better story quality.
In this paper, we introduce \textbf{\em \system}\footnote{\system is available at
https://dixitdemo.github.io/index.html . The web page will be responsive and
can be used via mobile devices upon demo.}, an \textit{interactive} visual
storytelling system.
The purpose of \system is to enable users to control the story generation process and
create their preferred stories.
This system is inspired by our observation that people often start their short
visual story with the key objects or actions in the pictures, and then build
connections between these terms to form the final story.
However, recent studies using deep neural networks (DNNs) are generally
designed in an end-to-end fashion without intermediate human-recognizable
representations~\cite{huang2016visual,wang2018nometics}.
This closed process prevents users from adjusting intermediate variables and
thus makes it impossible to create diverse stories from a given image sequence.
In addition, as story generation is a subjective process depending on one's
imagination, there is no ground truth given identical inputs, and
automatic metrics fail to evaluate the quality of the story with an absolute
score. In sum, we have the following four main issues:

\begin{figure*}[t]
    \centering
    \includegraphics[
    width=1\textwidth]{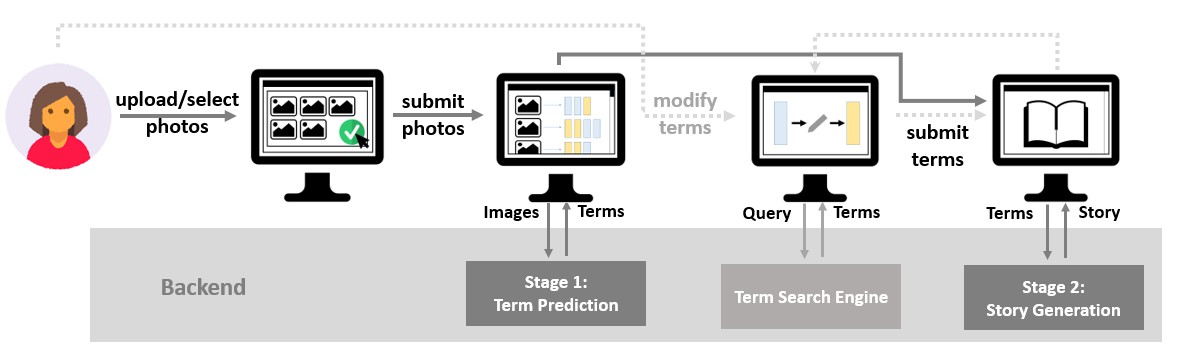}
    \caption{System architecture. The user initiates the process by uploading a sequence of photos. \system first extracts text \textit{terms} from each photo which describe the
objects or actions in the photo, and then allows the user to add new terms or remove existing terms. \system then generates a short story based on these terms.
}
    \label{fig:framework}
\end{figure*}

First, \textbf{lack of interpretability in intermediate DNN Layers.}
Because of the end-to-end DNN model's powerful ability as a universal
approximator, in recent years it has achieved significant improvements in many fields,
including computer vision and natural language processing.
However, though some studies have attempted to get inside the DNN black box, it
is difficult to clearly explain what information propagates to the next layer
in a DNN model.
For example, in visual storytelling, 
how exactly an end-to-end framework selects entities from images
remains a mystery    
and the model's behavior is difficult to interpret.
Second,
\textbf{inability to modify generated stories.}
With an end-to-end model, another problem is that the same input always yields
the same result. That is, we can obtain only a single story given
the same image sequence~-- but this goes against the creative nature of storytelling.
Although some generative models such as the variational autoencoder and the
generative adversarial network produce more results by sampling from a prior
distribution, they cannot generate storylines in the desired direction.
Third, \textbf{difficulty in automatic evaluation.}
Automatic metrics such as BLEU and METEOR have been shown to be insufficient to
evaluate the quality of a visual story generation
system~\cite{wang2018nometics}. Due to its subjective characteristics, human
evaluation, which takes money and time, has become an essential part of the
storytelling task.
Finally,
\textbf{data scarcity.} The quantity of image-story pair data is relatively
insufficient because visual storytelling is an complex task; thus data
annotation is expensive and time consuming.

To address these four issues, we introduce a two-stage interactive visual story
generation system with which (i) users can understand what model sees in the images as predicted terms represent the objects and actions in the images, and also observe how these terms influence story generation
, (ii) users can modify terms in the first
stage to control the story generation, (iii) we can obtain human evaluation by asking users to score the quality of the generated
story and the whole user-system interaction record in a turn is saved for future usage, and (iv) the modular nature of the
two-stage framework, which utilizes image caption data and story corpora
separately, allows us to train the visual storytelling model without
image-story pair data.

\section{Related Work}
Whereas end-to-end text generation from visual contents has been widely explored
for tasks such as image caption, video caption, and visual storytelling, recent work
has also been done on two-stage frameworks. 
Alexander et al. propose two-stage style transfer for image captioning.
They first extract objects and verbs from an image and generate a stylish
caption using an RNN trained on story corpora~\cite{Alexander2018Semstyle}. However,
each such caption is only a single story-like sentence and is independent of
other captions; put together, the captions do not constitute a context-coherent story. 
Xu et al. propose a skeleton-based narrative story generation method which
in the first stage extracts the skeleton of the input sentence after which they
generate the next sentence according to the given skeleton,
one-by-one~\cite{xu2018skeleton}. Although this framework shares with
our system the insight that key phrases are important for generating a coherent story, its
skeleton is extracted from pure text and does not serve as the grounding for
the story. Moreover, as 
the framework generates the skeleton after generating the sentence,  
it lacks the ability to consider all important terms
together to create the best fit.

\section{System}
\paragraph{Workflow.}
Users of our system either upload images from their devices or choose from
an image pool. 
After five images are uploaded or chosen, the first stage uses a term prediction
model that captures terms~-- i.e., the objects and actions in images~-- to help
users understand what the model sees in the current photos. As explained in 
Section~\ref{frame}, the actions are represented by verb frames.
In the second stage, the predicted terms are then concatenated and fed into a 
story generator to construct a story with appropriate and coherent context.
Note that the terms can be modified arbitrarily to control
the story generation. If users want a story that depicts objects other than
those detected from the images, they can remove the terms and add desired terms;
to this end the system also includes a term search function.
After the story is generated, the users are asked to rate the story; the
complete interaction record (including image selection and term modification)
is saved in the database. 
We expect the routine of modifying terms and generating a story to be repeated
many times, as the user controls the story generation process and gradually 
gets better at creating good stories. For instance, in Table~\ref{table:exmaple_compare_story},
the generated story in the second row is not perfect because some
predicted terms are, although reasonable, not preferred. However, after removing
these terms and adding better ones, users create a better story. 
This auto-labeling is an efficient way to provide samples for
the training of future story models.

\begin{figure}[b]
    \centering
    \frame{\includegraphics[
    width=1\columnwidth]{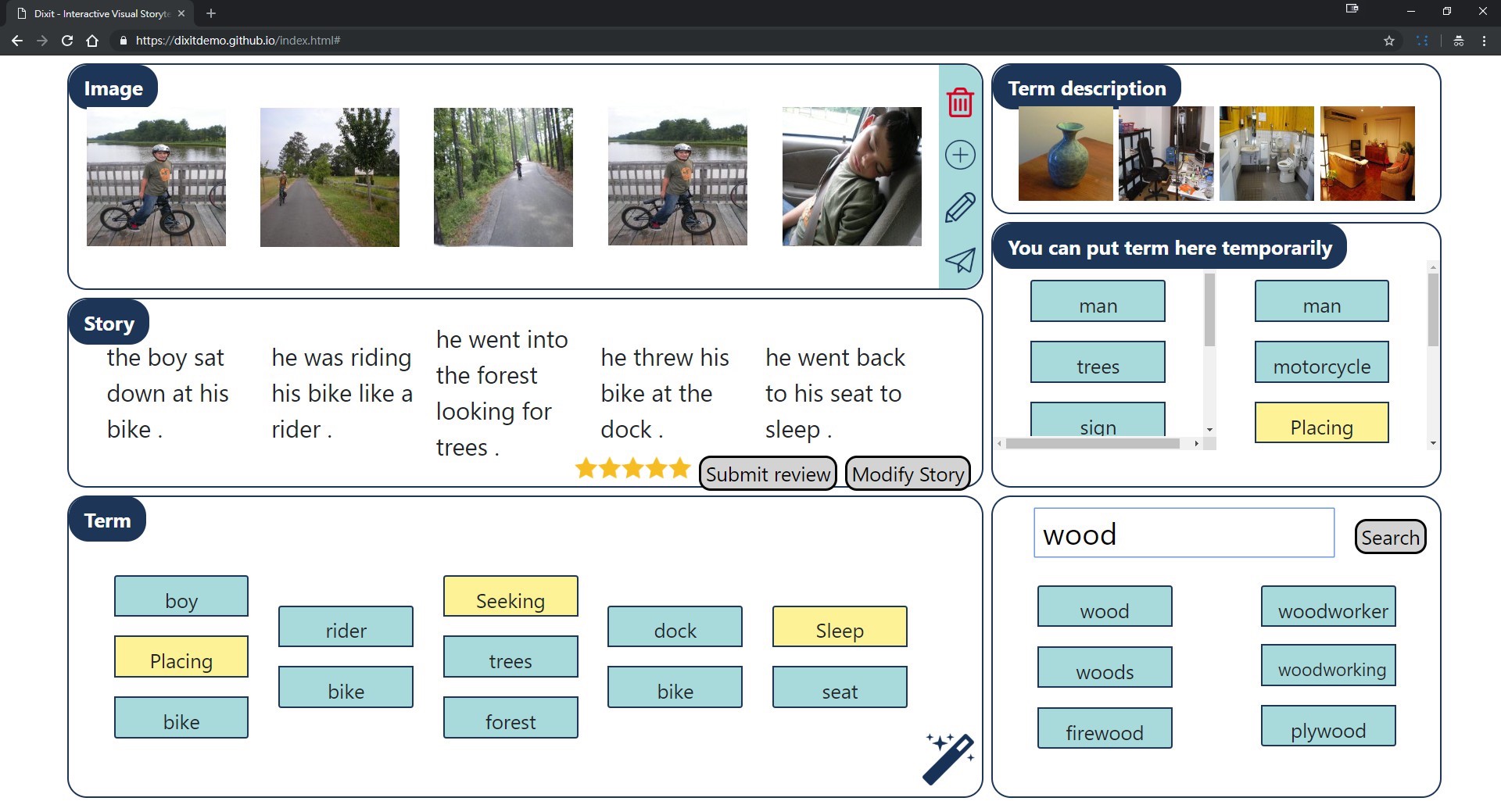}}

    \caption{System user interface}

    \label{fig:UI}
\end{figure}

\begin{table*}[!ht]
\centering
\resizebox{1\textwidth}{!}{
\begin{tabular}{llllll}
\toprule
 & \includegraphics[width=0.18\textwidth]{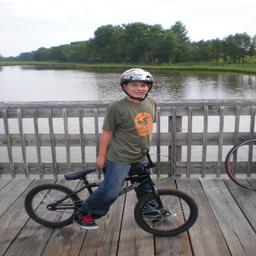}
 & \includegraphics[width=0.18\textwidth]{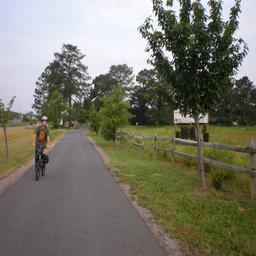}
 & \includegraphics[width=0.18\textwidth]{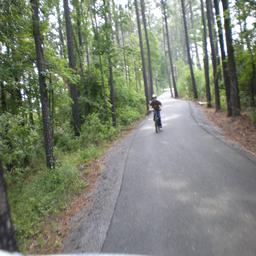}
 & \includegraphics[width=0.18\textwidth]{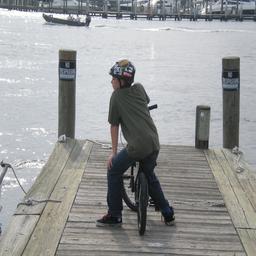}
 & \includegraphics[width=0.18\textwidth]{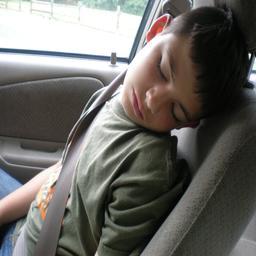}\\ \midrule
\textbf{Predicted Terms} & "man", "Placing",  "bike" & \begin{tabular}[c]{@{}l@{}}"motorcycle", "rider", \\ "Emptying", "bike"\end{tabular} & \begin{tabular}[c]{@{}l@{}}"trees", "sign", \\ "Preventing\_or\_letting",  \end{tabular}& \begin{tabular}[c]{@{}l@{}}"man", "Placing", "bench"\end{tabular} & \begin{tabular}[c]{@{}l@{}}"boy", "seat", "Placing" \end{tabular} \\ \midrule
\textbf{Story} & 
\begin{tabular}[c]
{@{}l@{}} the man was sitting \\on his bike.\end{tabular} & \begin{tabular}[c]{@{}l@{}}he was riding his bike \\ with a rider on his \\motorcycle.\end{tabular} & \begin{tabular}[c]{@{}l@{}}he stopped at a stop\\sign from a trees.\end{tabular} & \begin{tabular}[c]{@{}l@{}}the man sat on\\the bench.\end{tabular} & \begin{tabular}[c]{@{}l@{}}the boy sat in his seat.\end{tabular} \\\midrule\midrule
\textbf{Modified Terms} & "boy", "Placing", "bike" & \begin{tabular}[c]{@{}l@{}}"rider", "bike"\end{tabular} & \begin{tabular}[c]{@{}l@{}}"Seeking", "trees",  "forest"\end{tabular} & \begin{tabular}[c]{@{}l@{}}"bike", "dock"\end{tabular} & \begin{tabular}[c]{@{}l@{}}"Sleep", "seat"\end{tabular} \\ \midrule
\textbf{New Story} & \begin{tabular}[c]{@{}l@{}}the boy sat down at\\ his bike.\end{tabular} & \begin{tabular}[c]{@{}l@{}}he was riding his bike \\ like a rider. \end{tabular} & \begin{tabular}[c]{@{}l@{}}he went into the forest \\ looking for trees. \end{tabular} & \begin{tabular}[c]{@{}l@{}}he threw his bike at\\ the dock.\end{tabular} & \begin{tabular}[c]{@{}l@{}}he went back to his seat \\ to sleep. \end{tabular} \\ \midrule
\end{tabular}
}
\caption{Via term adjustment, story generation is focused on specific objects or actions}
\label{table:exmaple_compare_story}
\end{table*}

\paragraph{User Interface.}
A screenshot of the user interface is shown in Figure~\ref{fig:UI}. The user's
work flow starts at the left-top block, where the user is asked to upload their
own images or choose from the image pool. After the users submit images to
the system, the predicted terms are shown in the left-bottom block. We
meanwhile provide a term searching area (right bottom) using which the user searches for and
modifies predicted terms. 
Term search is used for nouns and verb frames.
For nouns, the search function retrieves nouns 
containing the query sub-string.
For instance, for the query \textit{wood}, the search engine
returns \textit{wood}, \textit{woods}, \textit{firewood}, \textit{woodworker},
\textit{woodworking}, and \textit{plywood}. 
While searching for verb frames, 
instead of directly querying by the frame name,
we search for lexical units of the verb frame. Results for the
query \textit{watch} would include \textit{seeking}, of which
\textit{watch} is a lexical unit of \textit{seeking}.
After clicking the magic wand icon in the left-bottom area, the selected terms
are sent to our story generation module, after which the generated story is retrieved
and shown in the left middle area. The user is then asked to rate the story via the
star icons. Since the concept of terms can be unfamiliar to users, we
provide a term description area in the right-top corner. When user hovers the pointer
over a term, the description of that term is displayed in the term
description area. For noun terms, the descriptions are related images
illustrating that noun, whereas for verb terms, the descriptions are textual
explanations of the verbs.
In the middle, on the right, a storage area is provided for 
the user to temporarily store unused terms.

\begin{figure*}
    \centering
    \includegraphics[
    width=1\textwidth]{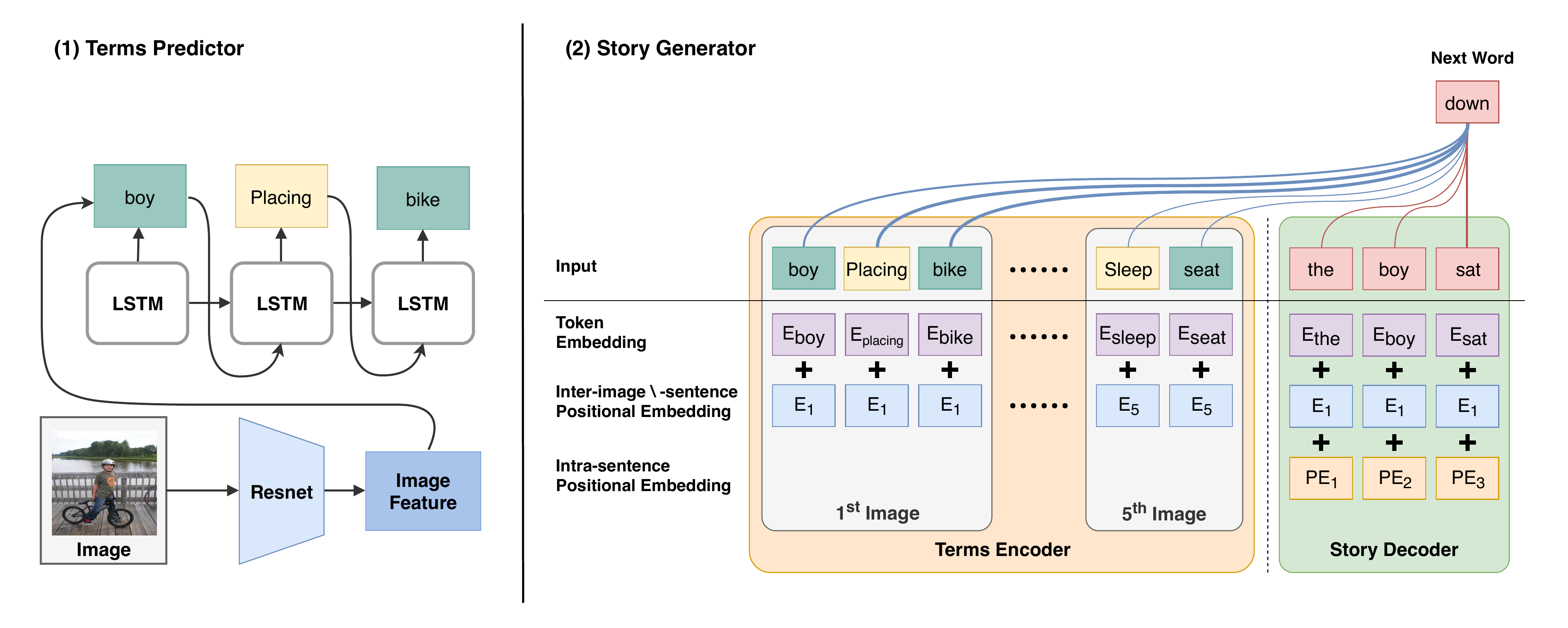}
	 \caption{Architecture of (1) term prediction model and (2) story
	 generation model. Predicted terms in green boxes denote noun terms,
	 and terms in yellow boxes denote verb frames. In the training process, which 
	 uses the a single vocabulary, we treat both noun terms and verb frames equally.
	 Term box colors are used here to emphasize their different origins.}

    \label{fig:transformer}
\end{figure*}

\section{Data Preparation}
\label{frame}
Due to the modular nature of the proposed system, we train the model in two stages4
separately, without parallel data.

To train the term prediction model, we
extract terms from captions in the COCO dataset~\cite{huang2016visual} as the
gold labels for the first stage to identity terms in images. The selection of
terms is inspired by Semstyle~\cite{Alexander2018Semstyle}. Each sentence is
represented as a combination of several noun (objects) and verb (actions) terms,
which preserves the most crucial information needed to generate stories. Given
a sentence, we extract verb terms by using Open-SESAME~\cite{swayamdipta17open}
to predict verb frames and leave nouns as noun terms. 
Frames are defined in FrameNet~\cite{baker1998berkeley}, a lexical database of
semantic frame. The basic idea of FrameNet is straightforward: each frame is a
description of a type of event, relation, or entity, and the participants in
it.

To train the story generator, we use ROCStories
Corpora~\cite{mostafazadeh2016roc}, which contains stories with five
sentences. For each story, we use a technique similar to back-translation to
build our term-to-story corpora. That is, we transform stories to their
corresponding terms, and then use these term-story pairs to train the story
generator.

\section{Image-to-Term Prediction}
As shown in Figure~\ref{fig:transformer}, the image-to-term prediction process
can be considered a simplified image captioning problem which instead predicts
the terms of an image. Given an image, we first convert it
to the RGB format and resize it to 3x224x224 (channel~* width~* height)
and then utilize a pre-trained Resnet-152~\cite{he2016deep} model to extract
image features. Resnet is a very deep neural network model with residual links
between layers, originally pre-trained for image classification. The
literature has suggested that it captures image features from
low-level components like arcs to high-level concepts.  
With the image features produced in the previous step, we feed image features
into an LSTM~\cite{hochreiter1997long} to generate terms.
Note that as described in Section ~\ref{frame}, the LSTM model is trained with
images and terms extracted from their corresponding captions in the COCO
dataset; this pipeline is similar to that depicted in~\cite{Alexander2018Semstyle}.

\section{Term-to-Story Generation}

Transformer~\cite{vaswani2017attention} is used as the basic model for 
story generation. In the Transformer paper, Vaswani et al. propose 
positional embedding as sequential information to replace the recurrent neural
network. Here, we extend Transformer to be sentence-aware
with the addition of an inter-sentence positional embedding to indicate the
order of sentences within a story. This is shown in
Figure~\ref{fig:transformer}. We do not consider the order of the input terms
because terms represent only what appears in the images and need no sequential
signal. At decoding time, an input embedding is the summation of the token
embedding, the inter-image/-sentence positional embedding (InterPE), and
the intra-sentence positional embedding (IntraPE). \\
In particular,
\begin{itemize}

\item The InterPE is randomly initialized and updated during the training
procedure. Similar to the segment embedding in BERT~\cite{devlin2018bert}, we
differentiate sentences/images by adding $E_{s}$ to token embeddings of the $s$-th
sentence/image; $s$ denotes the order of a sentence/image in a story. $s \in \{1,
2, 3, 4,5\}$.
\item For IntraPE, we follow the Transformer implementation:
\begin{align*}
\text{IntraPE}_{(\textit{pos},\textit{2i})} = sin(pos/10000^{2i/d_{\mathit{model}}}) \\
\text{IntraPE}_{(\textit{pos},\textit{2i+1})} = cos(pos/10000^{2i/d_{\mathit{model}}})
\end{align*}
where $pos$ is the position and $i$ is the dimension. $d_{\mathit{model}}$ denotes
the dimension of the input and output token.
The IntraPE parameters are fixed, and the sinusoidal representation allows the
model to extrapolate the sentence length that is longer than the training
instances. The IntraPE $PE_{s}$ is added to the $s$-th token embedding in a
sentence. $s \in \{1, 2,...,n\}$ and $n$ is the max sentence length. 

\end{itemize}

Based on the terms predicted for each image in the first step, we concatenate
terms predicted for the five images as the input to the story generator. While
decoding the story at every time step, the sentence-aware Transformer
generates the context-coherent story conditioned on all input terms and the
prior words of the story.

\section{Conclusion and Future Work}
In this paper, we propose a two-stage interactive visual storytelling system
that extracts terms (nouns and verb frames) and generates stories based on
these terms. Users interactively adjust the terms to produce the desired
story, and then record and rate the story quality for later model improvement. By
splitting the task into two steps and their middle semantic term layer, the
training data are not limited to the pair of images and their story, but
include more accessible images with captions, and the text stories themselves.
In addition, the process of visual story generation becomes interpretable. We
believe this is a worthy direction for the advance of visual storytelling.

\section*{Acknowledgement}
This research is partially supported by Ministry of Science and Technology, Taiwan, under Grant no. MOST108-2634-F-001-004-, MOST108-2634-F-002-008- and MOST107-2218-E-002-009-.

\bibliographystyle{ACM-Reference-Format}
\balance
\bibliography{main}


\begin{thebibliography}{11}


\ifx \showCODEN    \undefined \def \showCODEN     #1{\unskip}     \fi
\ifx \showDOI      \undefined \def \showDOI       #1{#1}\fi
\ifx \showISBNx    \undefined \def \showISBNx     #1{\unskip}     \fi
\ifx \showISBNxiii \undefined \def \showISBNxiii  #1{\unskip}     \fi
\ifx \showISSN     \undefined \def \showISSN      #1{\unskip}     \fi
\ifx \showLCCN     \undefined \def \showLCCN      #1{\unskip}     \fi
\ifx \shownote     \undefined \def \shownote      #1{#1}          \fi
\ifx \showarticletitle \undefined \def \showarticletitle #1{#1}   \fi
\ifx \showURL      \undefined \def \showURL       {\relax}        \fi
\providecommand\bibfield[2]{#2}
\providecommand\bibinfo[2]{#2}
\providecommand\natexlab[1]{#1}
\providecommand\showeprint[2][]{arXiv:#2}

\bibitem[\protect\citeauthoryear{Baker, Fillmore, and Lowe}{Baker
  et~al\mbox{.}}{1998}]%
        {baker1998berkeley}
\bibfield{author}{\bibinfo{person}{Collin~F Baker}, \bibinfo{person}{Charles~J
  Fillmore}, {and} \bibinfo{person}{John~B Lowe}.}
  \bibinfo{year}{1998}\natexlab{}.
\newblock \showarticletitle{The berkeley framenet project}. In
  \bibinfo{booktitle}{\emph{Proceedings of the 17th international conference on
  Computational linguistics-Volume 1}}. Association for Computational
  Linguistics, \bibinfo{pages}{86--90}.
\newblock


\bibitem[\protect\citeauthoryear{Devlin, Chang, Lee, and Toutanova}{Devlin
  et~al\mbox{.}}{2018}]%
        {devlin2018bert}
\bibfield{author}{\bibinfo{person}{Jacob Devlin}, \bibinfo{person}{Ming-Wei
  Chang}, \bibinfo{person}{Kenton Lee}, {and} \bibinfo{person}{Kristina
  Toutanova}.} \bibinfo{year}{2018}\natexlab{}.
\newblock \showarticletitle{Bert: Pre-training of deep bidirectional
  transformers for language understanding}.
\newblock \bibinfo{journal}{\emph{arXiv preprint arXiv:1810.04805}}
  (\bibinfo{year}{2018}).
\newblock


\bibitem[\protect\citeauthoryear{He, Zhang, Ren, and Sun}{He
  et~al\mbox{.}}{2016}]%
        {he2016deep}
\bibfield{author}{\bibinfo{person}{Kaiming He}, \bibinfo{person}{Xiangyu
  Zhang}, \bibinfo{person}{Shaoqing Ren}, {and} \bibinfo{person}{Jian Sun}.}
  \bibinfo{year}{2016}\natexlab{}.
\newblock \showarticletitle{Deep residual learning for image recognition}. In
  \bibinfo{booktitle}{\emph{Proceedings of the IEEE conference on computer
  vision and pattern recognition}}. \bibinfo{pages}{770--778}.
\newblock


\bibitem[\protect\citeauthoryear{Hochreiter and Schmidhuber}{Hochreiter and
  Schmidhuber}{1997}]%
        {hochreiter1997long}
\bibfield{author}{\bibinfo{person}{Sepp Hochreiter} {and}
  \bibinfo{person}{J{\"u}rgen Schmidhuber}.} \bibinfo{year}{1997}\natexlab{}.
\newblock \showarticletitle{Long short-term memory}.
\newblock \bibinfo{journal}{\emph{Neural computation}} \bibinfo{volume}{9},
  \bibinfo{number}{8} (\bibinfo{year}{1997}), \bibinfo{pages}{1735--1780}.
\newblock


\bibitem[\protect\citeauthoryear{Huang, Ferraro, Mostafazadeh, Misra, Devlin,
  Agrawal, Girshick, He, Kohli, Batra, et~al\mbox{.}}{Huang
  et~al\mbox{.}}{2016}]%
        {huang2016visual}
\bibfield{author}{\bibinfo{person}{Ting-Hao~K. Huang}, \bibinfo{person}{Francis
  Ferraro}, \bibinfo{person}{Nasrin Mostafazadeh}, \bibinfo{person}{Ishan
  Misra}, \bibinfo{person}{Jacob Devlin}, \bibinfo{person}{Aishwarya Agrawal},
  \bibinfo{person}{Ross Girshick}, \bibinfo{person}{Xiaodong He},
  \bibinfo{person}{Pushmeet Kohli}, \bibinfo{person}{Dhruv Batra},
  {et~al\mbox{.}}} \bibinfo{year}{2016}\natexlab{}.
\newblock \showarticletitle{Visual Storytelling}. In
  \bibinfo{booktitle}{\emph{Proceedings of NAACL 2016}}.
\newblock


\bibitem[\protect\citeauthoryear{Mathews, Xie, and He}{Mathews
  et~al\mbox{.}}{2018}]%
        {Alexander2018Semstyle}
\bibfield{author}{\bibinfo{person}{Alexander~Patrick Mathews},
  \bibinfo{person}{Lexing Xie}, {and} \bibinfo{person}{Xuming He}.}
  \bibinfo{year}{2018}\natexlab{}.
\newblock \showarticletitle{SemStyle: Learning to Generate Stylised Image
  Captions using Unaligned Text}.
\newblock \bibinfo{journal}{\emph{CoRR}}  \bibinfo{volume}{abs/1805.07030}
  (\bibinfo{year}{2018}).
\newblock


\bibitem[\protect\citeauthoryear{Mostafazadeh, Chambers, He, Parikh, Batra,
  Vanderwende, Kohli, and Allen}{Mostafazadeh et~al\mbox{.}}{2016}]%
        {mostafazadeh2016roc}
\bibfield{author}{\bibinfo{person}{Nasrin Mostafazadeh},
  \bibinfo{person}{Nathanael Chambers}, \bibinfo{person}{Xiaodong He},
  \bibinfo{person}{Devi Parikh}, \bibinfo{person}{Dhruv Batra},
  \bibinfo{person}{Lucy Vanderwende}, \bibinfo{person}{Pushmeet Kohli}, {and}
  \bibinfo{person}{James~F. Allen}.} \bibinfo{year}{2016}\natexlab{}.
\newblock \showarticletitle{A Corpus and Cloze Evaluation for Deeper
  Understanding of Commonsense Stories}. In
  \bibinfo{booktitle}{\emph{Proceedings of NAACL 2016}}.
\newblock


\bibitem[\protect\citeauthoryear{Swayamdipta, Thomson, Dyer, and
  Smith}{Swayamdipta et~al\mbox{.}}{2017}]%
        {swayamdipta17open}
\bibfield{author}{\bibinfo{person}{Swabha Swayamdipta}, \bibinfo{person}{Sam
  Thomson}, \bibinfo{person}{Chris Dyer}, {and} \bibinfo{person}{Noah~A.
  Smith}.} \bibinfo{year}{2017}\natexlab{}.
\newblock \showarticletitle{{Frame-Semantic Parsing with Softmax-Margin
  Segmental RNNs and a Syntactic Scaffold}}.
\newblock \bibinfo{journal}{\emph{arXiv preprint arXiv:1706.09528}}
  (\bibinfo{year}{2017}).
\newblock


\bibitem[\protect\citeauthoryear{Vaswani, Shazeer, Parmar, Uszkoreit, Jones,
  Gomez, Kaiser, and Polosukhin}{Vaswani et~al\mbox{.}}{2017}]%
        {vaswani2017attention}
\bibfield{author}{\bibinfo{person}{Ashish Vaswani}, \bibinfo{person}{Noam
  Shazeer}, \bibinfo{person}{Niki Parmar}, \bibinfo{person}{Jakob Uszkoreit},
  \bibinfo{person}{Llion Jones}, \bibinfo{person}{Aidan~N Gomez},
  \bibinfo{person}{{\L}ukasz Kaiser}, {and} \bibinfo{person}{Illia
  Polosukhin}.} \bibinfo{year}{2017}\natexlab{}.
\newblock \showarticletitle{Attention is all you need}. In
  \bibinfo{booktitle}{\emph{Advances in Neural Information Processing
  Systems}}. \bibinfo{pages}{5998--6008}.
\newblock


\bibitem[\protect\citeauthoryear{Wang, Wang, Chen, and Wang}{Wang
  et~al\mbox{.}}{2018}]%
        {wang2018nometics}
\bibfield{author}{\bibinfo{person}{William~Yang Wang}, \bibinfo{person}{Xin
  Wang}, \bibinfo{person}{Wenhu Chen}, {and} \bibinfo{person}{Yuan{-}Fang
  Wang}.} \bibinfo{year}{2018}\natexlab{}.
\newblock \showarticletitle{No Metrics Are Perfect: Adversarial Reward Learning
  for Visual Storytelling}. In \bibinfo{booktitle}{\emph{Proceedings of NAACL
  2018}}.
\newblock


\bibitem[\protect\citeauthoryear{Xu, Zhang, Zeng, Ren, Cai, and Sun}{Xu
  et~al\mbox{.}}{2018}]%
        {xu2018skeleton}
\bibfield{author}{\bibinfo{person}{Jingjing Xu}, \bibinfo{person}{Yi Zhang},
  \bibinfo{person}{Qi Zeng}, \bibinfo{person}{Xuancheng Ren},
  \bibinfo{person}{Xiaoyan Cai}, {and} \bibinfo{person}{Xu Sun}.}
  \bibinfo{year}{2018}\natexlab{}.
\newblock \showarticletitle{A skeleton-based model for promoting coherence
  among sentences in narrative story generation}.
\newblock \bibinfo{journal}{\emph{arXiv preprint arXiv:1808.06945}}
  (\bibinfo{year}{2018}).
\newblock


\end{thebibliography}

\end{document}